\pdfoutput=1

\documentclass[11pt]{article}

\usepackage[final]{acl}

\usepackage{times}
\usepackage{latexsym}

\usepackage{mathtools}
\usepackage[T1]{fontenc}

\usepackage[utf8]{inputenc}
\usepackage{amssymb}
\usepackage{microtype}
\usepackage{float}
\usepackage{inconsolata}

\usepackage{graphicx}
\usepackage{multirow}
\usepackage{booktabs}
\usepackage[authormarkup=none,final]{changes}
\usepackage[]{todonotes}
\usepackage[normalem]{ulem}
\useunder{\uline}{\ul}{}

\definecolor{featBlue}{RGB}{51, 102, 204}
\definecolor{ruleBrown}{RGB}{102, 51, 00}
\newcommand{\our}{\textsc{ARISE}}

\newcommand{\bfl}{\mathbf l}
\newcommand{\bfx}{\mathbf x}

\newcommand{\Lcal}{\mathcal{L}}

\newcommand{\Ycal}{\mathcal{Y}}

\definecolor{teal}{rgb}{0.0, 0.5, 0.5}
\definecolor{turq}{rgb}{0.68, 0.93, 0.93}

\definechangesauthor[name={Amrith Krishna}, color=teal]{AK}
\definechangesauthor[name={Ganesh}, color=turq]{GR}

\title{\our: Iterative Rule Induction and Synthetic Data Generation for Text Classification}

\author{
Yaswanth M$^{\heartsuit0}$,
Vaibhav Singh$^{\spadesuit0}$,
\textbf{Ayush Maheshwari$^{\diamondsuit}$\thanks{Work done while at IIT Bombay, with BharatGen.}}, \\
\textbf{Amrith Krishna}$^{\clubsuit}$,
\textbf{Ganesh Ramakrishnan}$^{\clubsuit\spadesuit}$ \\
$^{\spadesuit}$Indian Institute of Technology Bombay,
$^{\clubsuit}$BharatGen\\
$^{\heartsuit}$Accenture, $^{\diamondsuit}$NVIDIA \\
}

\begin{document}
\maketitle
\begin{abstract}

We propose \our, a framework that iteratively induces rules and generates synthetic data for text classification. We combine synthetic data generation and automatic rule induction, via bootstrapping, to iteratively filter the generated rules and data. We induce rules via inductive generalisation of syntactic n-grams, enabling us to capture a complementary source of supervision. These rules alone lead to performance gains in both, in-context learning (ICL) and fine-tuning (FT) settings. Similarly, use of augmented data from ARISE alone improves the  performance for a model, outperforming configurations that rely on complex methods like contrastive learning.  Further, our extensive experiments on various datasets covering three full-shot, eight few-shot and seven multilingual variant settings demonstrate that the rules and data we generate lead to performance improvements across these diverse domains and languages.



\end{abstract}

\section{Introduction}

\footnotetext[0]{ Correspondence to: Yaswanth M <yaswanthm03@gmail.com> and
Vaibhav Singh <singhvaibhav@cse.iitb.ac.in>}
Large language models (LLMs) have facilitated the generation of high-quality synthetic data that often supplement available training data \cite{lin-etal-2023-selective} or even surpass crowd-sourced annotations \cite{augmentationPNAS,alizadeh2023open}. However, concerns of limited variance in such exemplars, leading to model collapse \cite{shumailov2023curse} or the failure to capture the tail of the true underlying distribution \cite{ding2024data}, remain. Similarly, forming multiple views of the available data by inducing rules, as a complementary source of supervision has shown to benefit various NLP tasks, including text classification \cite{maheshwari2021semi,dong-etal-2022-syntactic}.  In this work, we propose \our, a bootstrapping approach to iteratively refine synthetically generated exemplars and automatically induced rules, resulting in high quality entries with respect to a given classification task \cite{yarowsky-1995-unsupervised,varma2018snuba}. 

\begin{figure}[h] 
   \centering
   \includegraphics[trim={0 1.1cm 0 1cm},clip,scale=0.75]{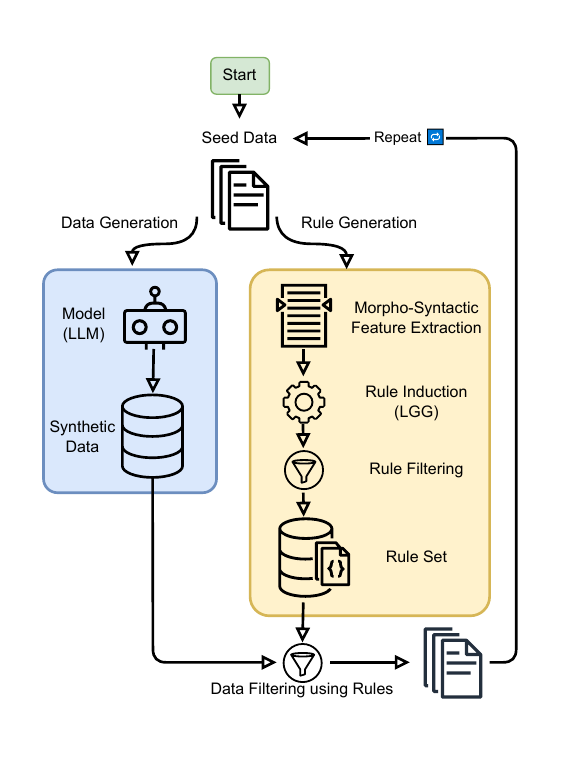}
   \caption{Overview of ARISE (\textbf{A}utomatic \textbf{R}ule \textbf{I}nduction using \textbf{S}yntactic tree g\textbf{E}neralization).} 
   \label{archi}
\end{figure}

Figure \ref{archi} provides an overview of \our. We start by using available training data as our seed. Using LLMs, we leverage in-context learning (ICL), with the seed as input to synthetically generate candidate exemplars \cite{liu-etal-2022-makes}. Similarly, we generate rule candidates, via inductive generalisation using least general generalization (LGG)~\cite{plotkin1971further,Raza_Gulwani_Milic-Frayling_2014} by extracting syntactic n-grams from the seed. Further, the induced rules are then filtered using a submodular graph cut-based function \cite{bajpai-etal-2024-fair, kothawade2021prism}. The exemplars and the rules we generate are task-specific and each exemplar and rule is associated with a label. Newly generated exemplars are filtered using rules that are generated from the already validated seed. These filtered exemplars are then added to the seed for the next iteration. Iteratively, we induce rules from synthetically generated data and use the induced rules for data filtering.

In \our, we boost supervision signals in two ways. With synthetic data generation we supplement the available training data \cite{DBLP:journals/corr/abs-1711-10160ratner,pryzant-etal-2022-automatic}. First, with rule induction, we obtain complementary signals that need not be explicitly captured from the existing data \cite{maheshwari2021semi, singhal-etal-2023-intendd}. Second, our rules are induced as generalized syntactic n-grams. Here, we aim to potentially capture morpho-syntactic information from the data, a view of data that need not be explicitly captured by state-of-the-art (SotA) systems in use. A classical NLP pipeline typically represents a string at multiple levels of abstraction which includes Part-of-Speech (PoS) tags, syntactic relations, \emph{etc.} \cite{manning-etal-2014-stanford}. \our~uses higher-order dependency structures as features and generalizes over these features using inductive generalization \cite{popplestone1970experiment} to induce the rules as generalized syntactic n-grams. 
 

We find applicability of both the rules and exemplars from \our, with consistent performance gains in various text classification setups. Specifically, we experiment with ICL and fine-tuning setups. In ICL, we focus on long-context ICL \cite{li2024long,bertsch2024context} and use the generated data as a pool from which exemplars are retrieved. Further, we incorporate our rules as explanations to the input and the exemplars. Similarly, we use the data for fine-tuning models, which include pre-trained LLMs, Qwen \cite{Qwen2,Qwen2.5} and RoBERTa \cite{liu2019roberta}. 

We perform extensive experiments on multiple text classification datasets, which include three full-shot, and eight few-shot datasets from the {\sc FewMany} benchmark \cite{yehudai2024llms}. Further, we perform multilingual experiments on seven languages using the {\sc MASSIVE} \cite{fitzgerald2022massive1mexamplemultilingualnatural} dataset. 


Our major contributions are as follows:

\begin{itemize}
    \item Use of rules and data from \our~results in statistically significant gains in all the experimental setups, as compared to the corresponding configuration without resources from \our. Specifically, we obtain state of the art (SotA) results in our full-shot and few-shot experiments when using \our.

    \item The rules we generate are shown to be effective, both during ICL and fine-tuning. Further, using the rules as explanations under ICL for CDR dataset results in SotA results. Similarly, fine-tuning Qwen jointly with data and the augmented rules from \our~has shown statistically significant improvements for Qwen and RoBERTa based models.

    \item Use of augmented data for few-shot setups in the {\sc FewMany} benchmark demonstrate the quality of the augmented data we produce. We show that simply using additional data from \our, as low as 20-shot additional data per class, can result in improved performance than incorporating complex approaches such as contrastive representation learning into the training process. 
    

    \item Our extensive experiments show that \our~is generalizable across multiple domains and multiple languages. We report a 7.21\%  increase in performance, compared to the model without any resources from \our, averaged across seven different languages.    

    \item We show that leveraging syntactic information as weak supervision for rule induction, brings a complementary source of supervision, which otherwise need not be captured by using string level data directly (\S \ref{ruleImpact}).
    

\end{itemize}

\section{ARISE - Automatic Rule Induction Using Syntactic Tree Generalization}
\label{sec:genSpace}

\begin{figure*}[h] 
    \centering
   \includegraphics[width=\textwidth,trim={0cm 0cm 0 0cm},clip]{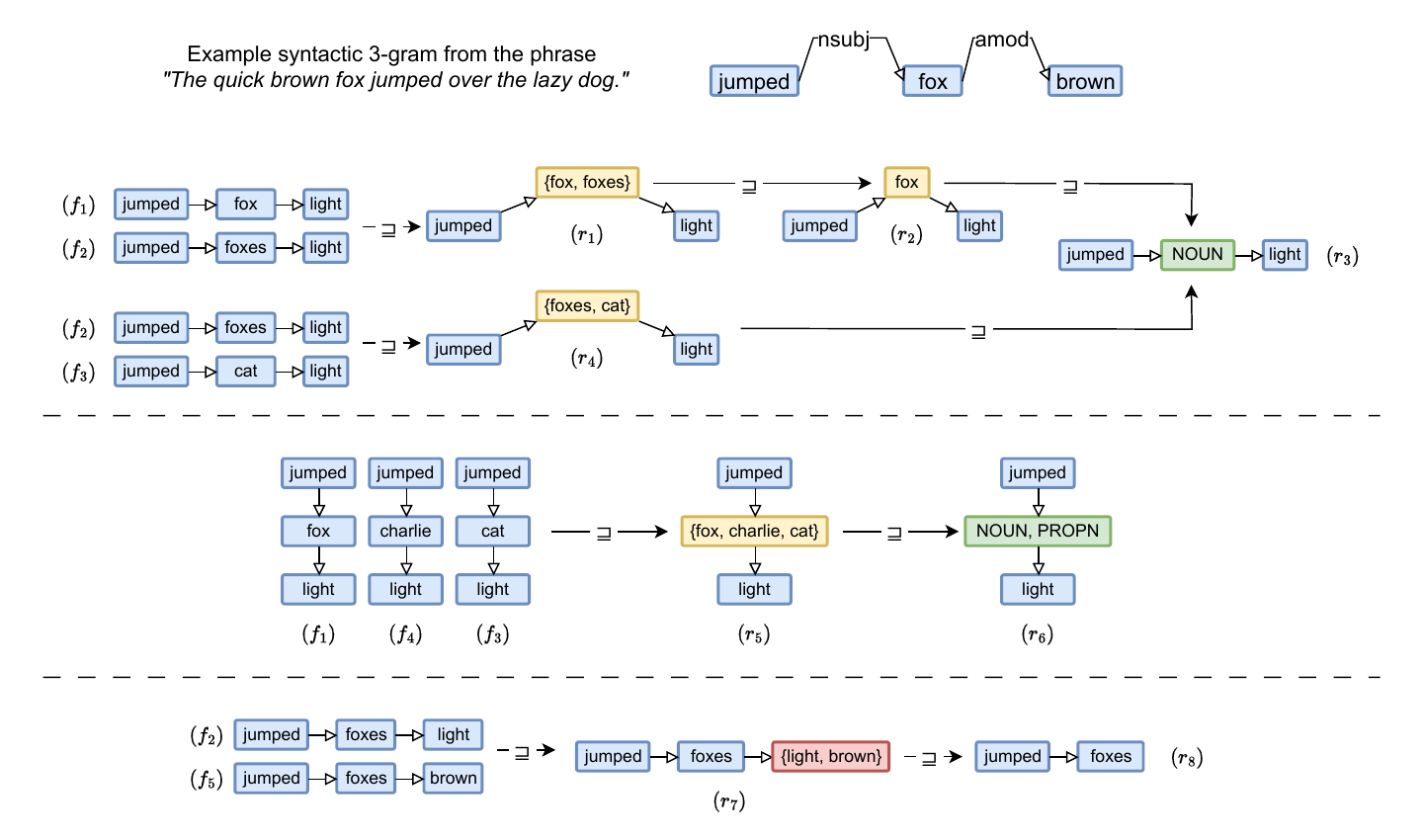}
   \caption{We induce rules via inductive generalization on syntactic n-grams, as shown (dependency relations omitted for brevety). The symbol `$\sqsupseteq$' denote a generalization operation. Trees labeled from $f_1$ to $f_5$ are instances of features. Similarly, trees labeled from $r_1$ to $r_8$ are rules.}
   \label{genSamples}
   \end{figure*}

Distributional hypothesis \cite{firth1957synopsis} is often realized using vector space models defined over a feature space \cite{turney2010frequency}. Inputs can be encoded as dense contextualized vectors \cite{peters-etal-2018-deep,devlin-etal-2019-bert} or  a sparse semantic space consisting of lexical n-grams, syntactic n-grams \cite{goldberg-orwant-2013-dataset}, higher order dependency features \cite{koo-collins-2010-efficient}, or even graph motifs \cite{biemann2016network}.

We induce rules that can capture complementary information that is not explicitly captured in pre-trained neural models. Hence, we focus on incorporating structured grammatical information typically used in a traditional NLP pipeline \cite{manning-etal-2014-stanford} such as Part-of-Speech (PoS) and syntactic information.  From dependency parses of input sentences, we extract induced subtrees as features. Each such feature is a syntactic n-gram, with the nodes as the words and the edges labeled with the dependency relations. We then induce rules via the inductive generalization of these features, using LGG \cite{Raza_Gulwani_Milic-Frayling_2014, thakoor2018multisynthesis}. 


For a text classification task with $k$ labels, we assume the  availability of a labeled dataset $\mathcal{D}$,  where $\mathcal{D} = \{(x_i,y_i)\}_{i=1}^{n}$, $x_i$ is an input document and $y_i \in \{l_1,l_2,...,l_k\}$ is a label. We obtain sentence-level dependency parses for each $x_i \in \mathcal{D}$. A feature space $\mathcal{F}_{t=1}^{f}$ is defined over higher-order factorization of the dependency parses in $\mathcal{D}$. Each feature $f_t \in \mathcal{F}$ is an induced subtree of the parses for sentences in $\mathcal{D}$. In Figure \ref{genSamples}, $f_1$ to $f_5$ denote instances of features in our feature space. These are syntactic n-grams extracted from  sentence-level dependency parses of the input. A feature covers a set of documents in which that feature occurs at least once.


Rules are generalizations of features. Now, $r_1$ to $r_8$, in Figure \ref{genSamples}, show various generalized rules induced from the features $f_1$ to $f_5$. If a generalized rule subsumes multiple features, then it covers a union of all the sets of documents corresponding to those features.   Our rules are induced as the least general generalization (LGG) over a set of features \cite{plotkinnote,plotkin1971further}. A feature can be a rule in itself, i.e. $\mathcal{F} \subseteq \mathcal{R}$, though it will be the most specific form of a rule. Previously, LGG was extensively used in information extraction \cite{califf-mooney-1997-relational,nagesh-etal-2012-towards}, program synthesis \cite{Raza_Gulwani_Milic-Frayling_2014,kitzelmann2010combined}, and in several other areas of relational learning \cite{muggleton1992efficient,zelle1994combining}.  

We define two forms of generalizations for forming the rules, both structural and linguistic. If rule $r_i$ is an induced subtree of $r_j$, then we can say that $r_i$ is more general than  $r_j$. Linguistic generalization include, substitution \cite{Raza_Gulwani_Milic-Frayling_2014,thakoor2018multisynthesis}, of the nodes containing words with their corresponding stems, and PoS tags \cite{galitsky2019least}. 

Figure \ref{genSamples} shows illustrative cases of generalization.  Let us consider a corpus from which features $f_1$ to $f_5$ are extracted.  Rules $r_1$ to $r_7$ show linguistic generalization. Similalry, rule $r_8$ shows structural generalization from $r_7$. Consider rules $r_1, r_4, r_5$ and $r_7$.  These rules contain nodes with a group of words. Similarly, $r_6$ represents a rule that has a group of PoS tags in one of the nodes. In linguistic generalization, multiple trees are generalized to a single tree by grouping words or PoS that differ in these individual trees. Here, $r_1$ is a generalisation of $f_1$ and $f_2$. Similarly, $r_4$ is a generalization of $f_2$ and $f_3$.

 We currently restrict the groupings at a node to be homogeneously typed, i.e. a set can either be that of inflected word forms, stems or of PoS tags, but not a mix of those. Further, the cardinality of such a group is set to an arbitrary upper bound, to avoid trivial generalisations. The rules we generate belong to $\mathcal{R}_{t=1}^{r}$. Here, for every input in $x_i \in \mathcal{D}$ it should either predict a label from $\{ 1, ..., k \}$, if the rule is applicable to the input. Else, it should abstain from making a prediction.

\subsection{Rule Induction via LGG}
\label{filtering}

We obtain features from dependency parses of the dataset $\mathcal{D}$. 
We consider only those subtrees that exactly have one of the six core dependency relations in them \cite{de-marneffe-etal-2014-universal, nivre-etal-2020-universal}. These core dependency relations are direct or indirect objects, nominal or clausal subjects, as well as clausal or open clausal complements. We partition the features into six mutually exclusive subsets, with each subset corresponding to one of the core relations. 

A complete lattice is constructed out of each partition, by adding a supremum and infimum element to the partition. Here, we add a rule `* $\xleftarrow{rel}$ {\sc *}', where `${rel}$' is the core-relation corresponding to the partition. It is the supremum for any element in the partition, as every element in the partition is subsumed by it and covers any document that has the relation present in it. We also define `$\epsilon$' as the infimum and it represents an empty rule that rules out any document in the input. The complete lattice provides a search space of rules over which the partial ordering is provided. Here, any two pair of subtrees have a least general generalization or a least upper bound \cite{Raedt2010}. In Figure \ref{genSamples}, $r_1$ is the LGG of $f_1$ and $f_2$.  $r_1$ represents all the sentences that either have $f_1$ or $f_2$ in their dependency parses. Similarly, $r_2$ and $r_3$ are also generalizations of $f_1$ and $f_2$, but not their LGG.


For every rule in the lattice, we compute its label-PMI vector, following \citet{singhal-etal-2023-intendd} and  \citet{Jin_Wanvarie_Le_2022}. Label-PMI vector is a vector of the pointwise mutual information scores of the rule corresponding to each label. From the vector, we consider its maximum score, denoted as {\sc L-PMI}. The label corresponding to {\sc L-PMI} is then assigned to the rule.  
From the lattice, we start bottom up and compute the LGG for every pair of rules. We induce the LGG as a rule, only if it has a higher {\sc L-PMI} than the individual rules in the pair. These induced rules form our candidate set of rules. For a given label $y_j$, the pointwise mutual information for the rule $r_t$ is given by,
$$\text{PMI}(y_j,r_t) = \frac{log |\mathcal{D}| \times \text{Count}(r_t,\mathcal{D}^{y_j})}{\text{Count}(r_t,\mathcal{D})*|\mathcal{D}^{y_j}|}$$

Here, $\text{Count}(a,b)$ implies the count of input documents having both $a$ and $b$. Similarly, $\mathcal{D}^{y_j}$ implies the set of  documents with the label $y_j$.   


Text classification tasks need not always have single sentence inputs. We generally assume a feature may appear in any of the sentences in the input, unless these sentences clearly have a predefined role in the task. For instance, interchanging the premise and hypothesis in natural language inference \cite{berant-etal-2011-global,DAGAN_DOLAN_MAGNINI_ROTH_2010} generally leads to different outcomes. In such cases, we induce rules for premise sentences and hypothesis sentences separately. Further, the {\sc L-PMI} is applied a second time, this time for a pair of rules, one induced from the premise and the other from the hypothesis. The 2-step  {\sc L-PMI} approach enables to reduce the combinatorial explosion which otherwise may happen, and is trivially extendable to tasks with multiple roles.  

\section{\our~Framework}
\label{sec:method}

\subsection{Rule Filtering}
\label{ruleFiltering}

We induce rules from a set of input documents (\S \ref{sec:genSpace}), which are expected to be noisy.  Two rules may even predict conflicting labels to a given input, akin to labeling functions in data programming \cite{DBLP:journals/corr/abs-1711-10160ratner,zhang2022survey}. Ideally, the final set of filtered rules needs to be accurate, diverse and high in coverage \cite{bajpai-etal-2024-fair}.

For rule filtering, we use the submodular graph-cut (GC) function \cite{kothawade2021prism}, as proposed by \citet{bajpai-etal-2024-fair}. Using GC, we select a final set of representative and diverse rules $\mathcal{R}_{f}$, from the set of candidate rules $\mathcal{R}$. For $\mathcal{R}_{f} \subseteq \mathcal{R}$, we define the GC function as:
$$ f_{GC}(\mathcal{R}_{f})=\sum_{r_{i} \in \mathcal{R}, r_{j} \in \mathcal{R}_{f}} s_{ij}-\lambda \sum_{r_{i},r_{j}\in \mathcal{R}_{f}}s_{ij} $$

Here, $\lambda \in [0,1]$ governs the diversity-representation trade-off, where higher $\lambda $ implies higher diversity in $\mathcal{R}_{f}$. $s_{ij}$ is the similarity score for rule pair $r_i$ and $r_j$. It is calculated as the weighted sum of the precision, coverage, and agreement between the pair of rules:
$$s_{ij} = \alpha(r_i)+\alpha(r_j)+w*\beta(\{r_i, r_j\}) + \gamma*\mu(r_i,r_j)$$

Here, $\alpha(r_i)=\text{Precision}(r_i)$, $\mu(r_i,r_j)$ is the agreement, calculated as the fraction of instances where both rules agree. $\beta(\{r_i, r_j\})$ is the coverage, calculated as the fraction of instances labeled by at least one of the rules.  

Our objective function is $\max_{|\mathcal{R}_{f}|\leq k} f_{GC}(\mathcal{R}_{f})$, where $k$ is a fixed budget \cite{kothawade2021prism}. We employ a greedy approach to choose a rule that maximizes the marginal utility 
$\textit{argmax}_{r_{i}\in\{\mathcal{R}-\mathcal{R}_{f}\}} f_{GC}(\mathcal{R}_{f}\cup \{r_i\})-f_{GC}(\mathcal{R}_{f})$. 
Please note that \citet{bajpai-etal-2024-fair} starts with an empty set, whereas we start with the existing rule set obtained from the previous round of bootstrapping. One round of filtering is completed until the fixed budget $k$ is exhausted.

\subsection{Bootstrapping Rules and Synthetic Data}

%

\citet{varma2018snuba} previously employed a bootstrapping based rule induction approach for labeling available unlabeled data. In \our, combining synthetic data generation and automated rule induction presents us with an opportunity to bootstrap and expand our labeled dataset and rules iteratively.


We start our bootstrapping with the training split of the available gold labeled data as the seed, as shown in Figure \ref{archi}. We synthetically generate new data for each class using prompt demonstrations, with the demonstrations retrieved from the seed set \cite{zhang-etal-2022-active, peng2024revisiting}. Similarly, we perform rule induction (\S \ref{sec:genSpace}) from the seed. The induced rule candidates are then filtered using the validation split of the available gold data and added to the rule set (Figure \ref{archi}). Similarly, data filtering for the synthetically generated exemplars is performed using the rule set. In data filtering, only those exemplars that match their generated label with the predicted label from the generative model are filtered. Finally, the seed set is expanded with the filtered data. The seed set and the rule set are expanded after every iteration of bootstraping.

\paragraph{Seed and validation set during few-shot: } \citet{zhu-etal-2023-weaker} observe that while weak supervision systems use limited training data, they heavily rely on the availability of a clean gold-labeled validation data for their performance gains. Hence, in our few-shot setups we do not use any validation split of the data and instead use only the few-shot training splits. Here, the gold labeled data, i.e. the few-shot training split,  is  used only as the validation data for rule filtering. Further, the initial iteration of synthetic generation happens in a zero-shot setup without demonstrations. Moreover, the rule candidates induced for the initial iteration is also from the synthetically generated samples, similar to subsequent iterations.





\paragraph{Data Augmentation:} We use prompt demonstration, long context ICL or few-shot depending on the task setup, to synthetically generate new labeled sentences using LLMs. For each label, we sample $m$ instances each of positive and negative samples from the seed set and then use it for generating new data samples \cite{10.1145/3617130smithLoop,lin-etal-2023-selective}. Our prompt demonstration approach includes label information, positive examples, and negative examples for synthetic generation. In addition to generating new data samples, we also perform paraphrasing of data samples in the seed set. By paraphrasing, we gain diverse syntactic structures for better rule induction.\footnote{For more details, refer \S \ref{paraphrase}}

\section{Experiments}

\paragraph{Dataset:}  We use three datasets, namely, {\sc Discovery} \cite{sileo-etal-2019-mining}, {\sc CDR} \cite{davis2017comparative,wrench} and {\sc ANLI} \cite{nie-etal-2020-adversarial}, as shown in Table \ref{tab:dataset_specs},  for our full-shot setup. Here, we use the full training split, unless hit by an upper bound of 15,000 labeled instances, when fine-tuning the models. For ANLI, we focus on the R3 Dataset. CDR is a binary true/false classification problem for a given document with
mentions of chemicals and diseases tagged. Similarly, {\sc ANLI} is a 3-class, natural language inference task. {\sc Discovery} attempts at identifying the appropriate discourse marker from a set of 174 classes for a given pair of statements.  For few-shot, we use the {\sc FewMany} Benchmark \cite{yehudai2024llms}, consisting of eight multiclass classification datasets. Here, we only use the 5-shot labeled data points from the training splits of the datasets involved. 
Finally, the multilingual experiments are performed using the {\sc MASSIVE} dataset \cite{fitzgerald2023massive} with intent classification as the task. Here, we use seven typologically diverse languages including Chinese, English, French, German, Hindi, Japanese, and Spanish. 
\begin{table}[h]
\centering
\setlength\tabcolsep{5pt} 
\footnotesize
\begin{tabular}{lcccc}
\toprule
\textbf{Dataset} & \textbf{Train} & \textbf{Dev} & \textbf{Test} & \textbf{\# Labels} \\ \midrule
{\sc Discovery}        & 1,566,000        & 87,000        & 87,000         & 174                  \\ 
{\sc ANLI}             & 100,459         & 1,200         & 1,200          & 3                    \\ 
{\sc CDR}              & 8,430              & 920            & 4,673             & 2                    \\
{\sc MASSIVE}          & 11,514          & 2,033         & 2,974          & 60                   \\ \bottomrule
\end{tabular}
\caption{Dataset statistics of the original datasets. for ANLI and {\sc Discovery}, we perform class-wise stratified sampling and do not use more than 15,000 labeled instances from the training split for fine-tuning setups. }
\label{tab:dataset_specs}
\end{table}


\paragraph{Data Generation:} We use GPT-3.5, GPT-4, and Claude 3 Opus for synthetic data generation. We generate label-specific data by prompt demonstration. Here, Using \citet{wu-etal-2023-openicl}, we perform $k$-NN retrieval from the seed data, with $k = min(n,150)$, where $n$ is the available data for a given label in the seed for positive demonstrations, and and equal amount of randomly sampled out of class samples as negative examples \cite{bertsch2024context,liu-etal-2022-makes}.  We use RoBERTa based sentence-embedding \cite{reimers-2019-sentence-bert} for sentence representation. For multilingual experiments, we experiment with \textit{direct} generation of the synthetic data in the target language, and also via \textit{translation} of synthetically generated English sentences. For translation, in addition to the three aforementioned LLMs we use NLLB-54B \cite{nllbteam2022language} and Google Translate. For translation in Hindi, we use \citet{gala2023indictrans}.

\subsection{Experimental Setup}

We incorporate the rules and exemplars from \our~in diverse text classification settings. 

\paragraph{In-context learning (ICL):} We experiment with three different configurations under long-context ICL using LLMs. One, is a \textbf{zero-shot} setup where we provide the input only with an explanation without any retrieved exemplars. Here, the explanations are obtained by phrasing the generated rules and their predictions as reasoning statements similar to the group of prompting techniques collectively referred to as thought generation prompting \cite{schulhoff2024prompt}. 
Two is the \textbf{\textit{k}-shot} setup where we add prompt demonstrations from the generated data into the prompt. Third is the \textbf{\textit{k}-shot-XP} setup where we append explanations in addition to the prompt demonstrations. 
For demonstrations, we provide rules leading to both correct and wrong label predictions for it, similar to that in Contrastive CoT \cite{chia2023contrastive}. We reuse the retrieval setup used for data generation. 
\paragraph{Fine-tuning (FT):}  We fine-tune an open-weight LLM \cite{Qwen2.5} and a smaller pre-trained LM \cite[PLM,][]{liu2019roberta, conneau-etal-2020-unsupervised}  in different configurations, under the \textbf{full-shot} setup. For fine-tuning the LLM, we employ PEFT using QLoRA. Our configurations for FT include; FT-base* where only the training data is used; FT-DA, where additional data from \our~are used; FT-J where only the rules corresponding to the training data are incorporated via Joint Learning using SPEAR \cite{maheshwari2021semi} and finally, FT-JDA, where both the data and rules from \our~is used. We also experiment with FT-JDX,  a variation where the rules are incorporated both as part of the input prompt and via Joint Learning.  

\paragraph{Joint Learning:} For incorporating the rules into our fine-tuning process, we follow SPEAR \cite{maheshwari2021semi}, a Joint Learning framework that learns over a feature-based classification model and a label aggregation (LA) model. The feature model is an LLM or a PLM and LA is a generative model \cite{cage}, learned via data programming, using the automatically induced rules as labeling functions.

LA is denoted as $P_{\theta}(\bfl_i, y)$, where $\bfl_i$ 
a vector that represents the firing of all LFs for an input $\bfx_i$. Each firing, $l_{ij}$ can be either 0 (abstain) or class label $k$ \cite{cage}.  
Our Joint Learning objective incorporates three different loss components for learning from labeled data. 
We provide a brief overview of each loss component below, while encouraging interested readers to \citet{maheshwari2022learning} for detailed information. The first component of the loss is the standard cross-entropy loss for the model $P_\phi^f$. The second component is the negative log-likelihood on the dataset. The third is the KL-Divergence between the predictions of the LA and $P_\phi^f$ models, which enforces consensus by aligning their predictions.
{\begin{align}\nonumber
\min_{\theta, \phi} &\sum_{i \in \Lcal} L_{CE}\left(P_\phi^f(y|\bfx_i), y_i\right)
 + LL_s(\theta| \Lcal)  
 \\ &+ \sum_{i \in \Lcal} KL\left( P_\phi^f(y|\bfx_i),P_\theta(y|\bfl_i)\right) \nonumber
\label{eq:objective}
\end{align}}

\paragraph{Base Models:} Our experiments are performed on one representative model for each of the following categories: (1) a closed-source LLM with access only via API, (2) an open-weight LLM, and (3) a pre-trained LM. Models like RoBERTa are still preferred in resource and latency conscious industry use cases, such as customer support, to larger models with few billion parameters. Factors include low latency, need for low hardware configuration and low cost, while still being competitive in several use cases. We choose GPT-4 Turbo \cite{openai2024gpt4technicalreport}, Qwen2.5-72B-Instruct \cite{Qwen2.5}, and RoBERTa-large \cite{liu2019roberta}, respectively, based on their category-wise performance on preliminary experiments using {\sc B77} and {\sc AP106} datasets. For multilingual setup, we employ XLM-RoBERTa \cite{conneau-etal-2020-unsupervised} based on observations from \cite{fitzgerald-etal-2022-massively}. Additionally, we utilize \citet[CPFT,][]{zhang-etal-2022-contrastive} for our contrastive learning \cite{pmlr-v119-chen20jsimclr} based baseline.



Accuracy is our primary evaluation metric. We use \citet{dozat2016deep}, a dependency parser,  to extract syntactic n-grams from input. We obtain induced subtrees of up to 3 nodes as rules. For few-shot, we perform all our experiments using 5 random splits and report the average \cite{yehudai2024llms}. For Joint Learning, we use 20\% of the synthetically generated data as the validation split, while using all the gold data as the training data. For learning the parameters for our rule filtering step (\S \ref{ruleFiltering}), we use the few-shot gold data as validation. For full-shot setups we use the standard train-validation-test splits.\footnote{For the prompt and hyperparameter details, refer: https://sites.google.com/view/ariserules/}  




\begin{table}[h]
\centering
\begin{tabular}{cllll}
\toprule
\multicolumn{2}{c}{Model Configuration}                                                                           & {\sc CDR}            & {\sc ANLI}           & {\sc DISC.}           \\
\midrule
\multicolumn{1}{c}{\multirow{3}{*}{\begin{tabular}[c]{@{}c@{}}GPT 4\\ (ICL)\end{tabular}}} & zero-shot & 85.89          & 79.53          & 2.34           \\  
\multicolumn{1}{c}{}                                                                     & k-shot    & 88.95          & 81.59          & 31.10         \\  
\multicolumn{1}{c}{}                                                                     & k-shot-XP & \textbf{92.13} & {\ul 86.78}    & 35.44          \\ \midrule
\multicolumn{1}{c}{\multirow{3}{*}{\begin{tabular}[c]{@{}c@{}}Qwen\\ (ICL)\end{tabular}}}  & zero-shot & 82.86          & 71.27          & 8.59           \\  
\multicolumn{1}{c}{}                                                                     & k-shot    & 84.24          & 73.19          & 39.70           \\  
\multicolumn{1}{c}{}                                                                     & k-shot-XP & 86.84          & 84.05          & 47.36          \\ \midrule
\multicolumn{1}{c}{\multirow{6}{*}{\begin{tabular}[c]{@{}c@{}}Qwen\\ (FT)\end{tabular}}}   & FT-base*  & 82.05          & 58.20          & 92.29          \\  
\multicolumn{1}{c}{}                                                                     & FT-J      & 85.27          & 63.08          & 92.70           \\  
\multicolumn{1}{c}{}                                                                     & FT-JXP     & 85.63          & 85.19          & 92.40           \\  
\multicolumn{1}{c}{}                                                                     & FT-DA     & 87.76          & 75.69         & 95.33 \\  
\multicolumn{1}{c}{}                                                                     & FT-JDA    & {\ul 90.16}    & 78.30           & {\ul 95.72} \\  
\multicolumn{1}{c}{}                                                                     & FT-JDX   & 90.08    & \textbf{88.37} & \textbf{95.81} \\ \midrule
\multicolumn{1}{c}{\multirow{6}{*}{\begin{tabular}[c]{@{}c@{}}PLM\\ (FT)\end{tabular}}}    & FT-base*  & 81.78          & 53.82          & 90.60           \\  
\multicolumn{1}{c}{}                                                                     & FT-J      & 84.72          & 57.78          & 90.88          \\  
\multicolumn{1}{c}{}                                                                     & FT-JXP     & 84.58          & 57.85          & 90.67          \\ 
\multicolumn{1}{c}{}                                                                     & FT-DA     & 86.94          & 62.35          & 93.02   \\ 
\multicolumn{1}{c}{}                                                                     & FT-JDA     & 86.61          & 62.87          & 93.26    \\ 
\multicolumn{1}{c}{}                                                                     & FT-JDX    & 86.74          & 62.95          & 93.43   \\ \bottomrule
\end{tabular}
\caption{Results in ICL and FT setups. Numbers in \textbf{boldface} and { \ul underline} represent best and the second-best configurations, respectively. Here, PLM refers to RoBERTa-large. FT-base* is the only configuration that does not incorporate \our.}
\label{tab:full-shot-results}
\end{table}

\section{Results}

\begin{table}[]
\centering
\begin{tabular}{clll}
\toprule
Dataset & Configuration & Vanilla & ARISE \\
\midrule
\multirow{4}{*}{\begin{tabular}[c]{@{}c@{}}CDR\end{tabular}} & zero-shot & 74.56          & \textbf{85.89} \\  
 &k-shot    & 83.43           & \textbf{88.95}         \\  
  &k-shot + Aug. & 83.56 &  \textbf{88.95}     \\
 &k-shot-XP + Aug. & 89.35 & \textbf{92.13}      \\ \midrule
\multirow{4}{*}{\begin{tabular}[c]{@{}c@{}}DISC.\end{tabular}} & zero-shot & 0.84          & \textbf{2.34}           \\  
 &k-shot    & 8.73           & \textbf{31.10}         \\  
  &k-shot + Aug. & 26.97 & \textbf{31.10}     \\
 &k-shot-XP + Aug. & 32.11 & \textbf{35.44}      \\ \bottomrule
\end{tabular}
\caption{ICL Experiments in GPT-4 that compares both ARISE, 
and ARISE-less scenarios under comparable conditions}
\label{tab:vanilla}
\end{table}

Use of data and rules from \our, results in statistically significant  gains for all the datasets, both under full-shot and few-shot setups, including multilingual few-shot scenario. In full-shot setup, we outperform SotA models for both CDR, Discovery and ANLI, with more than an 8\% \cite{zhao-etal-2024-pareto}, 7\%\footnote{Not a comparable model} \citep[MTL;][]{sileo-etal-2019-mining} and 18\% \cite{kavumba-etal-2023-prompting} increase, respectively.


Table \ref{tab:full-shot-results} shows the results for various configurations where data and rules from \our~are used. FT-base* is the only configuration where no information from \our~is used. FT-base* however is fine-tuned on the available training splits of the corresponding datasets. Qwen FT-JDX, the configuration that uses Joint Learning with rules, rules as explanations and augmented data reports the best results for ANLI and Discovery with an absolute gain of more than 30\% and 3\%, respectively. For Discovery, gains from rules are not significant, as Qwen FT-DA achieves comparable results to Qwen FT-JDX. Similarly,  GPT 4 k-shot-XP achieves the best results for CDR, outperforming even the fine-tuned version with an absolute gain of roughly 10\% as compared to Qwen FT-base*. Qwen FT-JDA and FT-JDX, both using Joint Learning with rules is the second best model. For CDR and ANLI, using both additional data and rules lead to statistically significant gains.\footnote{Statistical significance is performed by t-test ($p < 0.05$)}


 Qwen models benefit from Joint Learning for CDR and ANLI, even after when their results saturate with additional data (FT-JDA). Here, both additional data and rules have shown to benefit the models and bring in complementary supervision signals. However, Joint Learning does not lead to statistically significant gains, once fine-tuning is performed with additional data for RoBERTa (FT-DA vs. FT-JDA for RoBERTa) 


Within ICL, GPT 4 outperforms Qwen in both CDR and ANLI, but Qwen outperforms GPT 4 in Discovery. Discovery has a large label space of 174 labels, and these are common terms which are typically used as markers between two statements. Qwen being an open-weight model, we were able to constrain the output space using constrained decoding.  
While there exist similar approaches with structured output generation in GPT 4,  we have limited control with GPT 4 compared to an open-weight model with constrained decoding.

Table \ref{tab:vanilla} compares ICL results with both \our, and \our-less scenarios under comparable conditions for both CDR and {\sc Discovery} datasets. We observe gains with \our, for all the casess we compared. Here, \textit{Vanilla zero-shot} does not use rules as explanations for the input from  \our, whereas \textit{Vanilla k-shot} retrieve examples only from the training split and uses no augmented data at all. \textit{Vanilla k-shot + Aug.} uses data augmentation proposed by \citet{lin-etal-2023-selective}. Finally, \textit{k-shot-XP + Aug.} uses the above augmentation setting, but adds explanation from \our. We find consistent performance improvements in all the configurations when \our~components are increasingly used. 





\subsection{Impact of Rules}
\label{ruleImpact}
We previously claimed that we obtain complementary supervision signals with rules compared to a setup like FT-base*. We validate the claim and 
observe statistically significant gains from the rules in ICL for all the three datasets, and in two of three datasets except for Discovery in fine-tuning.


\paragraph{Rules used for Joint Learning:} FT-J and FT-JXP are two settings, in Table \ref{tab:full-shot-results}, which are trained jointly using the rules from \our, but only with the original training split. For CDR, FT-J results in a percentage increase of 3.92 and 3.59 for both Qwen and RoBERTa, respectively, compared FT-base*. Similarly for ANLI, we observe a percentage increase of 8.38 and 7.35, respectively, for Qwen and RoBERTa respectively. Moreover, ANLI reports statistically significant gains with Qwen for FT-JXP, XP implying rules used also as explanations, compared to FT-J. We did not see any additional gains with RoBERTa when adding rules as explanations. We hypothesize this is due to lack of instruction tuning. 

For other datasets, FT-JXP configuration does not lead to statistically significant gains. For Discovery, the gains in absolute terms were not statistically significant for both Qwen and RoBERTa. However, there was no performance degradation for this dataset. Summarily, we find overall gains in using Joint Learning while fine-tuning task-specific models.  Qwen and RoBERTa both show similar trends and comparable gains with Joint Learning as compared to simple fine-tuning.

\paragraph{Rules used as explanations:}  We use our rules, along with their label predictions, as an explanation for the input. k-shot-XP, uses a subset of exemplars comapred to k-shot, irrespective of the source pool from which it is retrieved. In spite of having lesser number of exemplars, adding contrastive explanations leads to  further gains in our experiments for both Qwen and GPT 4. Further, use of explanations in k-shot settings with GPT 4 led to the highest performance for CDR among configurations. Similarly, we report the second best performance for ANLI using GPT 4, which has a percentage increase of more than 16 from the previous SotA. Previous SotA was a fine-tuned model, with 1/3rd of total training data, while our configuration under discussion is purely under ICL.

\subsection{Impact of Generated Data} 

We find that using generated data, both for training and for ICL, leads to statistically significant gains in all configurations for the three datasets. In our experiments, we generate synthetic data in multiples of the original training data size. We generate data from 1$\times$ to 6$\times$ of the original data.

\paragraph{Fine-tuning:} For all the three datasets, adding additional data beyond the full training data  during fine-tuning leads to significant gains. For both discovery and CDR, we observe gains until 1.5$\times$ times more data is added to the training data. For ANLI, we observe gains by doubling the training data size, i.e. 1$\times$ the training data. We observe that the training saturates after when more data is being added to these datasets, until we tried with 3$\times$ more data. Our observations hold true for both Qwen and RoBERTA, where tried the fine-tuning. 

\paragraph{ICL:} There are more  exemplars than that can fit into the 128K context windows for all the three datasets. However, increasing the pool of available data leads to improved outcomes in ICL, with the help of retrieval. For ANLI and CDR, we consistently had less than 40\% presence of instances from the original training data for cases with 1$\times$ and above. With Discovery, we observe similar patterns but only after  1.5$\times$. We stop our experiments with 3$\times$ data as the overlap between retrieved ICL exemplars was more than 90\%  by then.




\subsection{Few-shot Setup}

Few-shot learning setups are particularly valuable in industry applications involving text classification tasks with a large number of classes ($>50$). The high annotation cost and resource demands in such settings can be mitigated by adopting few-shot strategies. Previous research has explored contrastive learning methods \cite{zhang-etal-2022-contrastive} for learning better semantic space for input representation, weak supervision techniques such as the joint learning framework proposed by \citet{maheshwari2021semi}, and synthetic data generation strategies \cite{lin-etal-2023-selective}. In this work, we evaluate the effectiveness of \our's data augmentation method in comparison to the other two approaches. Specifically, we examine the data augmentation thresholds at which the benefits of competing techniques become statistically insignificant.


We only use 5-shot gold data for each class, and augment data from 1$\times$ to 256$\times$ in multiples of 4 \cite{lin-etal-2023-selective}. 
For all datasets we find statistically significant gains to Joint Learning until 32$\times$ augmented data is used. However, with more supplementary data at 64$\times$ and beyond, we do not find statistical significance between models that use Joint Learning compared to the one not using Joint Learning.  The only exception in the benchmark here is Amazon products, for which we find statistically significant gains to Joint Learning even with 200$\times$ data, but the gains disappears at 256$\times$ data. Similarly gains from contrastive learning in our CPFT baseline starts to disappear with 25-shot data (4$\times$ augmented) itself for all the datasets. 

Moreover, we observe that fine-tuned variants of RoBERTa and Qwen models report comparable performances and do not have any statistical significance between their results. RoBERTa and Qwen report an average accuracy of 94.18 and 95.04 respectively. Here, the only dataset with a difference in statistical significance between these two are Amazon Products. 62.07\% and 67.84\% respectively are the accuracy for RoBERTa and Qwen for Amazon Products. While Qwen-ICL performs worse than Qwen-FT with an average accuracy of 90.38\%, GPT 4 reports scores similar to the fine-tuned variants 95.46\%.

\paragraph{Multilingual Experiments:} 
On an average \our~reports an absolute improvement of 7.21\% points compared to the base model, on the 5-shot gold and $128x$ augemented data per class. The results show that our approach is applicable across a typologically diverse set of languages. We find \textit{translation} of synthetically generated English sentences leads to empirically better results as compared to \textit{direct} generation of data in the target language. 
The latter approach results in an absolute  drop of 1.27\% points. 
Moreover, GPT 4, under ICL reports an average of 84.15\% accuracy as compared to the 80.4\% accuracy reported by the RoBERTa Model. 

\section{Conclusion}

We propose \our, a framework that iteratively generates and refines both synthetic data and rules. Overall we find gains in using our rules and data in both ICL and FT for more than 15 datasets we consider. Further, \our~outperforms strong competitive baselines under comparable conditions. We also show the effectiveness of combining diverse sources of supervision that enable incorporating complementary and supplementary information beyond the available gold data to achieve SotA results.







\section*{Limitations}

A major challenge with \our, currently is the overhead with the rule induction. We currently use syntactic n-grams with upto 3 nodes as our features. The search space exponentially increases as the nodes of the subtree increase, limiting our ability to induce higher-order tree structures as rules. While we currently rely on labeled instances of synthetically generated data, a strength of weak supervision is to incorporate unlabeled data. Several real-world scenarios often come up where unlabeled data is readily available. It needs to be further investigated whether synthetically generated labeled data can match the quality of real-world unlabeled data in the context of weak supervision. The current work does not explore this line of work, though it appears to be an important question that requires further investigation.

\section*{Ethics Statement}
All experiments conducted in this study utilize publicly
available datasets. We use publicly hosted APIs of GPT and Claude for synthetic data generation. The prompts used in this study included guardrails in the form of instructions to avoid generating problematic content.

\section*{Acknowledgements}
We acknowledge BharatGen and IIT Bombay for providing resources and support to Vaibhav Singh. Yashwanth M. acknowledges Accenture for their support during this research. We also extend our appreciation to the reviewers for their valuable feedback.



\appendix

\section{Appendix}
\label{sec:appendix}
\subsection{Related Work}

\our~uses syntactic n-grams as its rules. Use of syntactic contexts in constructing feature space for downstream NLP tasks has been extensively explored in several of the past works \cite{liang-etal-2011-learning, goldberg-orwant-2013-dataset,biemann2016network}. \citet{goldberg-orwant-2013-dataset} released a large scale collection of syntactic n-grams obtained from 3.4 million books. \citet{biemann2016network} looks from a network science perspective and focuses on graph motifs. Learning feature functions using morphosyntactic information as horn clauses has shown to benefit under a low-resource setting for languages such as Czech and Sanskrit, often requiring less than 10\% of labeled training data required for neural counterparts \cite{10.1162/coli_a_00390,krishna-etal-2018-free}. 

Using syntactic context, we incorporate signals that may not otherwise be explicitly captured in large language models. Further, we automate the generation and filtering of such rules by relying extensively on rule induction approaches \cite{varma2018snuba,bajpai-etal-2024-fair,lao2010relational}. Additionally, we consider our rule generation approach as a restricted instance of program synthesis via least general generalization as demonstrated in \citet{Raza_Gulwani_Milic-Frayling_2014}, and \citet{thakoor2018multisynthesis}.

Data augmentation and generation in text has become effortless with LLMs \cite{ding2024data}. However, that does not ensure obtaining data with relevant supervisory signals, highlighting the need for targeted data filtering or generation \cite{pmlr-v139-killamsetty21a,mirzasoleiman2020coresets}. This may include data scoring and ranking \cite{lin-etal-2023-selective}, iterative data generation \cite{rao-etal-2023-makes}, bootstrapping \cite{varma2018snuba} or targeted subset selection \cite{pmlr-v37-wei15}. \citet{wang-etal-2023-lets} and \citet{lee-etal-2024-llm2llm} explore similar themes by utilizing errors from language models to iteratively refine a synthetic training dataset. Similarly, \cite{hoang-etal-2018-iterative} discussed back-translation in the context of machine translation to augment training data. In \our, we use bootstrapping approach for data filtering and apply our filtering on synthetically generated data, instead of unlabeled data from an existing corpus.

\subsection{Joint Learning with Rules}
\label{contrastive}

The few-shot classifier is trained using SPEAR \cite{maheshwari2021semi}, a Joint Learning framework that learns a feature-based classification model and a label aggregation (LA) model. The feature model is a pre-trained neural network and LA is a generative model \cite{cage}, learned via PWS, using the automatically induced rules as labeling functions. Formally, LA is denoted as $P_{\theta}(\bfl_i, y)$, where $\bfl_i$ 
a vector that represents the firing of all LFs for an input $\bfx_i$. Each firing, $l_{ij}$ can be either 0 (abstain) or class label $k$ \cite{cage}.  %
The model learns $K$ parameters $\theta_{j1}, \theta_{j2}, \ldots, \theta_{jK}$ for each class corresponding to each LF $l_{j}$.
\begin{equation}\displaystyle P_{\theta}(\bfl_i, y) = \frac{1}{Z_\theta} \prod_{j=1}^m \psi_\theta(l_{ij}, y)\label{eq:joint1}\end{equation}\begin{equation}\psi_{\theta}(l_{ij},y) = \begin{cases}\exp(\theta_{jy})  & \text{if $l_{ij}\ne 0$} \\1 & \text{otherwise.}\end{cases}\label{eq:decoupledthetas}\end{equation}

\begin{equation}\begin{split}Z_\theta = & \sum_y \prod_j\sum_{l \in \{1, 0\}} \psi_\theta(l, y)
\\        = & \sum_{y\in \Ycal}\prod_j(1+\exp(\theta_{jy}))\end{split}{}\end{equation}


Following \citet{maheshwari2021semi}, our Joint Learning objective incorporates three different loss components for learning from labeled data. 
We provide a brief overview of each loss component below, while encouraging interested readers to \cite{maheshwari2021semi} for detailed information.
{\begin{align}\nonumber
\min_{\theta, \phi} &\sum_{i \in \Lcal} L_{CE}\left(P_\phi^f(y|\bfx_i), y_i\right)
 + LL_s(\theta| \Lcal)  
 \\ &+ \sum_{i \in \Lcal} KL\left( P_\phi^f(y|\bfx_i),P_\theta(y|\bfl_i)\right) \nonumber
\label{eq:objective}
\end{align}}
 
The first component of the loss is the standard cross-entropy loss for the model $P_\phi^f$. The second component is the negative log-likelihood on the dataset. The third is the KL-Divergence between the predictions from LA and $P_\phi^f$, which enforces consensus by aligning their predictions. 

\subsection{Paraphrasing for Diverse Rules}
\label{paraphrase}

We employ various techniques to generate diverse syntactic structures for our pool of available features to be used in the rule induction stage. First, we perform active to passive voice sentence phrasing and vice versa using LLMs. Second, we perform dependency tree morphing \cite{sahin-steedman-2018-data}, to obtain simplified morphed  dependency trees. Here, we remove peripheral relations like adjectives, such that the core semantics of the sentence is still preserved. Third, we apply role prompting \cite{schulhoff2024prompt}, by prompting LLMs to rewrite sentences in the style of well-known authors \cite{wikipediaCategory21stcenturyEnglish}. Role prompting was exclusively applied during monolingual experiments.

\end{document}